\documentclass[10pt,twocolumn,letterpaper]{article}

\usepackage[pagenumbers]{iccv} % To force page numbers, e.g. for an arXiv version

\definecolor{iccvblue}{rgb}{0.21,0.49,0.74}
\usepackage[pagebackref,breaklinks,colorlinks,allcolors=iccvblue]{hyperref}
\usepackage{bbding}

\def\paperID{15547} % *** Enter the Paper ID here
\def\confName{ICCV}
\def\confYear{2025}

\title{CCL-LGS: Contrastive Codebook Learning for 3D Language Gaussian Splatting}

\author{
Lei~Tian$^{1}$, \quad
Xiaomin~Li$^{1}$, \quad
Liqian~Ma$^{2}$, \quad
Hao~Yin$^{1}$, \quad
Zirui~Zheng$^{1}$, \\
Hefei~Huang$^{1}$, \quad
Taiqing~Li$^{1}$, \quad
Huchuan~Lu$^{1}$, \quad
Xu~Jia$^{1,}$\textsuperscript{\Envelope} %\Envelope
\\
$^1$Dalian University of Technology \;\;
$^2$ZMO AI Inc.\\
}

\begin{document}

\twocolumn[{%
\vspace{-0.4cm}
\renewcommand\twocolumn[1][]{#1}%
\maketitle
\begin{center}
    \centering
    \captionsetup{type=figure}
    \includegraphics[width=\textwidth]{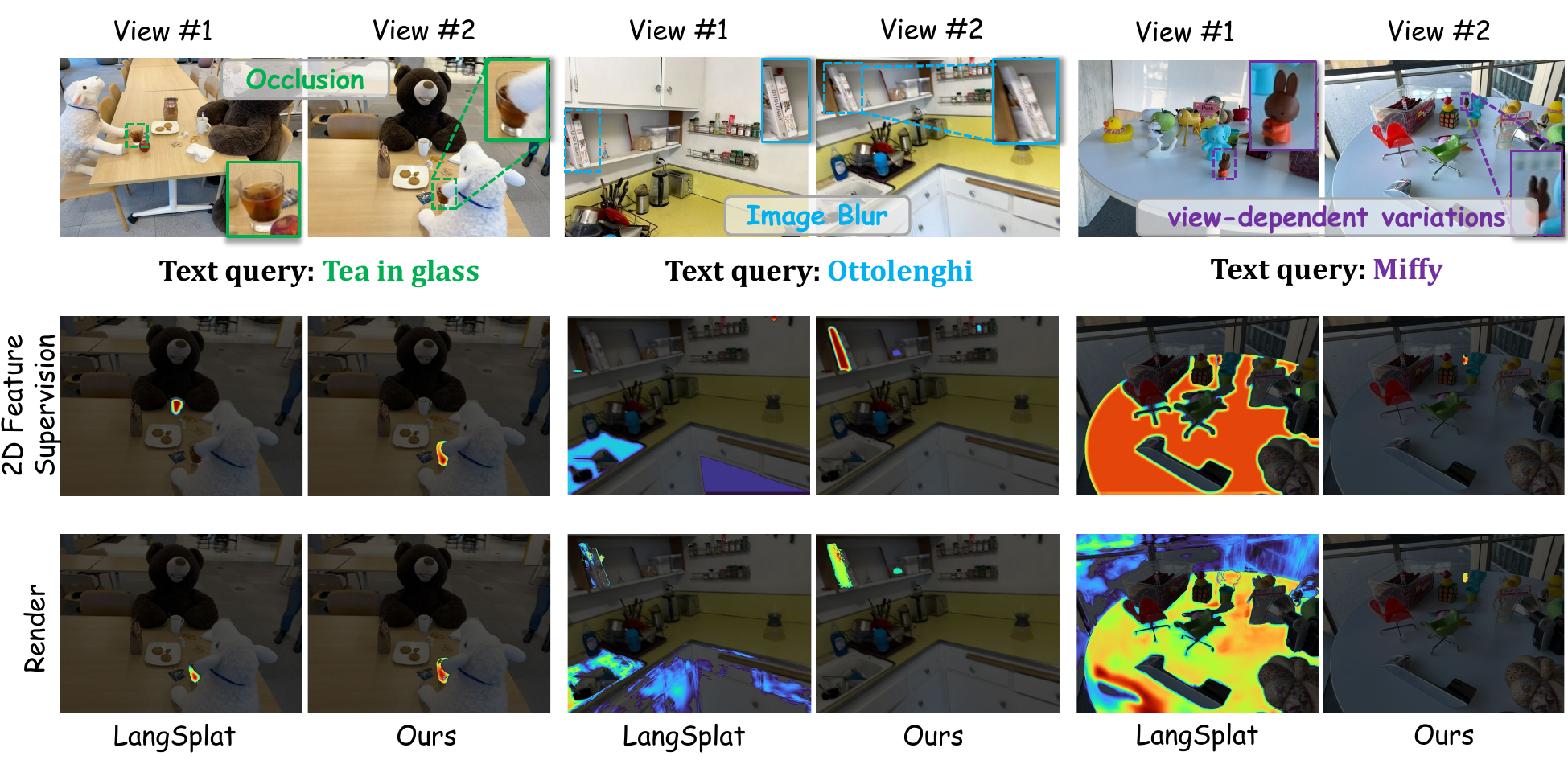}
    \captionof{figure}{Quantitative comparison of our method and LangSplat under three challenging scenarios: Occlusion, Image Blur, and View-Dependent Variations. The results clearly demonstrate the superior performance of our approach, which exhibits greater robustness and fidelity in handling these challenges compared to LangSplat.}
    \label{fig:teaser}
    
\vspace{-6pt}
\end{center}
}]
{\let\thefootnote\relax\footnotetext{{\textsuperscript{\Envelope}Corresponding authors.}}}

\begin{abstract}
Recent advances in 3D reconstruction techniques and vision-language models have fueled significant progress in 3D semantic understanding, a capability critical to robotics, autonomous driving, and virtual/augmented reality. However, methods that rely on 2D priors are prone to a critical challenge: cross-view semantic inconsistencies induced by occlusion, image blur, and view-dependent variations. These inconsistencies, when propagated via projection supervision, deteriorate the quality of 3D Gaussian semantic fields and introduce artifacts in the rendered outputs. To mitigate this limitation, we propose CCL-LGS, a novel framework that enforces view-consistent semantic supervision by integrating multi-view semantic cues. Specifically, our approach first employs a zero-shot tracker to align a set of SAM-generated 2D masks and reliably identify their corresponding categories. Next, we utilize CLIP to extract robust semantic encodings across views. Finally, our Contrastive Codebook Learning (CCL) module distills discriminative semantic features by enforcing intra-class compactness and inter-class distinctiveness. In contrast to previous methods that directly apply CLIP to imperfect masks, our framework explicitly resolves semantic conflicts while preserving category discriminability. Extensive experiments demonstrate that CCL-LGS outperforms previous state-of-the-art methods.
Our project page is available at \url{https://epsilontl.github.io/CCL-LGS/}.
\end{abstract}    
\section{Introduction}
\label{sec:intro}

Significant progress has been made in 3D reconstruction, particularly with the advent of 3D Gaussian splatting (3DGS)~\cite{3dgsrealtime}, which enables the generation of high-fidelity 3D color representations and supports real-time rendering of novel viewpoints. Meanwhile, vision-language models such as CLIP~\cite{clip} and LSeg~\cite{lseg} have bridged the gap between the two modalities, enabling the generation of rich and dense semantic maps for images without the need for additional supervision. With such advancing technologies, 3D semantic understanding, which aims to obtain 3D semantic representations from multi-view images and corresponding camera poses, has made rapid progress. This task has a wide range of applications, including robotics~\cite{Rl-gsbridge}, autonomous driving~\cite{autosplat}, and VR/AR~\cite{advancing}.

This task is particularly challenging due to factors such as semantic ambiguity (e.g., a point on the tip of the nose may correspond to both ``nose" and ``face" queries) and the long-tail distribution of vocabulary queries. Early methods like LERF~\cite{lerf} tackle these issues by extending NeRF~\cite{nerf} with multi-scale CLIP features and dynamic 2D mapping selection, achieving accurate semantic localization. However, their reliance on exhaustive multi-scale rendering leads to inefficiency, and patch-based feature extraction often fails to capture precise object boundaries, resulting in scale misalignment and performance degradation. Building upon the rise of explicit 3D representations such as 3DGS~\cite{3dgsrealtime}, recent works~\cite{langsplat, goi, 3dvlgs} differ from earlier NeRF-based methods by leveraging visual foundation models to extract dense, pixel-level semantic features that guide the construction of 3D semantic fields.
While differing in implementation, these methods generally follow a common paradigm: supervising 3D semantic representations by projecting them to 2D views and comparing the rendered results with features extracted from pre-trained vision-language models.

However, this paradigm based on 2D priors relies heavily on the assumption that semantic supervision remains consistent across different views. In practice, as illustrated in Fig.~\ref{fig:teaser}, factors such as occlusion, motion blur, and view-dependent variations can introduce significant inconsistencies in the 2D semantic features of the same object across viewpoints. Recent methods~\cite{langsplat, legaussians, goi, 3dvlgs} built upon 3DGS~\cite{3dgsrealtime} reconstruct a 3D semantic field from 2D features and leverage 3D geometric consistency to partially address cross-view semantic inconsistencies. However, since they still rely on 2D supervision, significant inconsistencies in the input features can propagate into the 3D space. This makes it difficult to maintain semantic coherence across views and often leads to artifacts in the rendered novel views. While many existing approaches enforce geometric constraints to enhance multi-view consistency, they often overlook the explicit modeling of 2D feature alignment across views—an underexplored yet critical factor that limits semantic reconstruction performance in open-world scenarios.

In this paper, we propose CCL-LGS, a novel framework for view-consistent 3D semantic reconstruction. Our key innovation lies in establishing view-consistent semantic supervision through a specially designed three-stage pipeline, enabling the reconstruction of a 3D Gaussian semantic field. Specifically, we first extract accurate instance masks using SAM~\cite{sam}, then align cross-view correspondences via zero-shot tracking, and finally distill semantics through a Contrastive Codebook Learning (CCL) module. The proposed CCL module introduces contrastive metric learning to enforce intra-class feature compactness while maintaining inter-class feature distinctiveness. This design effectively mitigates semantic ambiguities introduced by incomplete or noisy masks. Unlike prior approaches that directly apply CLIP to imperfect masks, our framework not only establishes reliable semantic correspondences across views but also preserves category-wise distinctiveness, leading to more robust and consistent 3D semantic reconstruction. Owing to its proficiency in 3D open-vocabulary scene understanding, our method could benefit a variety of downstream applications.
The main contributions of our work can be summarized as follows:
\begin{itemize}
\item We propose a novel framework, CCL-LGS, which integrates view-consistent semantic supervision to enable the reconstruction of 3D Gaussian semantic fields.
\item We develop a Contrastive Codebook Learning (CCL) module that resolves semantic ambiguities by enforcing intra-class compactness and inter-class distinctiveness, enabling robust semantic representation even with noisy or incomplete masks.
\item Extensive experiments on benchmark datasets demonstrate that our approach achieves state-of-the-art performance in open-vocabulary semantic segmentation tasks.
\end{itemize}
\section{Related Work}
\textbf{3D Neural Representations.} Recent advances in 3D scene representation have yielded impressive results, particularly with neural radiation fields (NeRF)~\cite{nerf}, which has excelled in novel view synthesis by generating highly realistic new perspectives. However, NeRF’s reliance on implicit neural networks to model entire scenes results in lengthy training and rendering times.
In contrast, explicit and hybrid scene representation approaches~\cite{cole2021differentiable, dai2023hybrid, prokudin2023dynamic, pointnerf, ripnerf} typically employ techniques such as hash grids and point clouds to alleviate the computational burden associated with large neural networks. Notably, 3D Gaussian Splatting (3DGS)~\cite{3dgsrealtime} has emerged as a promising alternative by representing scenes as collections of 3D Gaussian ellipsoids that can be efficiently rasterized into images. This method not only accelerates the rendering process but also allows for intuitive manipulation of individual scene components. Motivated by the success of 3D Gaussian Splatting in novel view synthesis, many studies have extended its application to other domains~\cite{text23d, dreamgaussian, gs-slam, gaussianflow, gart} to fully leverage its efficient rendering capabilities. In this paper, we utilize 3D Gaussian Splatting~\cite{3dgsrealtime} for 3D neural representations.

\noindent
\textbf{Visual Foundation Models.} Recent advances in visual foundation models have redefined computer vision by leveraging massive datasets and high-capacity architectures to tackle a wide range of tasks. Models such as CLIP~\cite{clip} employ contrastive learning to integrate visual and textual modalities, while the Segment Anything Model (SAM)~\cite{sam} delivers impressive zero-shot performance by generating high-quality segmentation masks across diverse scenarios. Self-supervised models like DINO~\cite{dino} and DINOv2~\cite{dinov2} further enrich this landscape by extracting fine-grained semantic features, including object boundaries and scene layouts, which are critical for downstream applications.
In parallel, advances in video object segmentation have opened new avenues for extending these robust visual representations into the temporal domain. Recent video segmentation methods focus on maintaining consistent object masks across frames, thus capturing dynamic temporal cues. For instance, XMem~\cite{xmem} leverages zero-shot tracking strategies, employing a memory mechanism to robustly segment objects in videos despite challenges like occlusion or appearance variations. SAM~\cite{sam} is an image segmentation model designed for single images, while SAM2~\cite{sam2} builds upon its capabilities to enable video object segmentation. By taking a video sequence and reference frame masks as input, SAM2 aims to address the challenge of long-term tracking by consistently predicting object masks in any target frame across extended video sequences. Therefore, in this paper, we employ SAM and CLIP to extract ground-truth features for baseline comparisons, using SAM2 for mask matching, ensuring a fair evaluation.

\noindent
\textbf{3D Scene Understanding.} Earlier works, such as Semantic NeRF~\cite{inplacescene} and Panoptic NeRF~\cite{panopticnerf}, pioneered the integration of 2D semantic or panoptic annotations into 3D radiance field representations, enabling zero-shot scene understanding. Building on this foundation, subsequent studies~\cite{decompnerf, neural3f} explored the use of pixel-aligned semantic features directly lifted into 3D, thereby moving beyond reliance on predetermined semantic labels. More recent methods for scene understanding focus on developing multi-modal 3D representations from posed images, supporting novel viewpoint rendering for tasks such as open-vocabulary semantic segmentation. For NeRF-based techniques, LERF~\cite{lerf} integrates a language field into NeRF by leveraging spatial positions and physical scales to produce CLIP vectors. Among 3DGS-based methods, LangSplat~\cite{langsplat} utilizes an autoencoder to perform dimensionality reduction on multi-scale features for extracting 2D semantic features. LEGaussians~\cite{legaussians} and GOI~\cite{goi} classify the extracted features using a codebook-based approach. Building upon the features extracted by LangSplat, 3D VL-GS~\cite{3dvlgs} places greater emphasis on language features by incorporating data enrichment and a cross-modal rasterizer. Although these methods adopt different technical strategies, they all share a reliance on projection-based supervision and overlook the issue of semantic inconsistency.

\section{Method}
\begin{figure*}[t]
\centering
\includegraphics[width=1.0\textwidth]{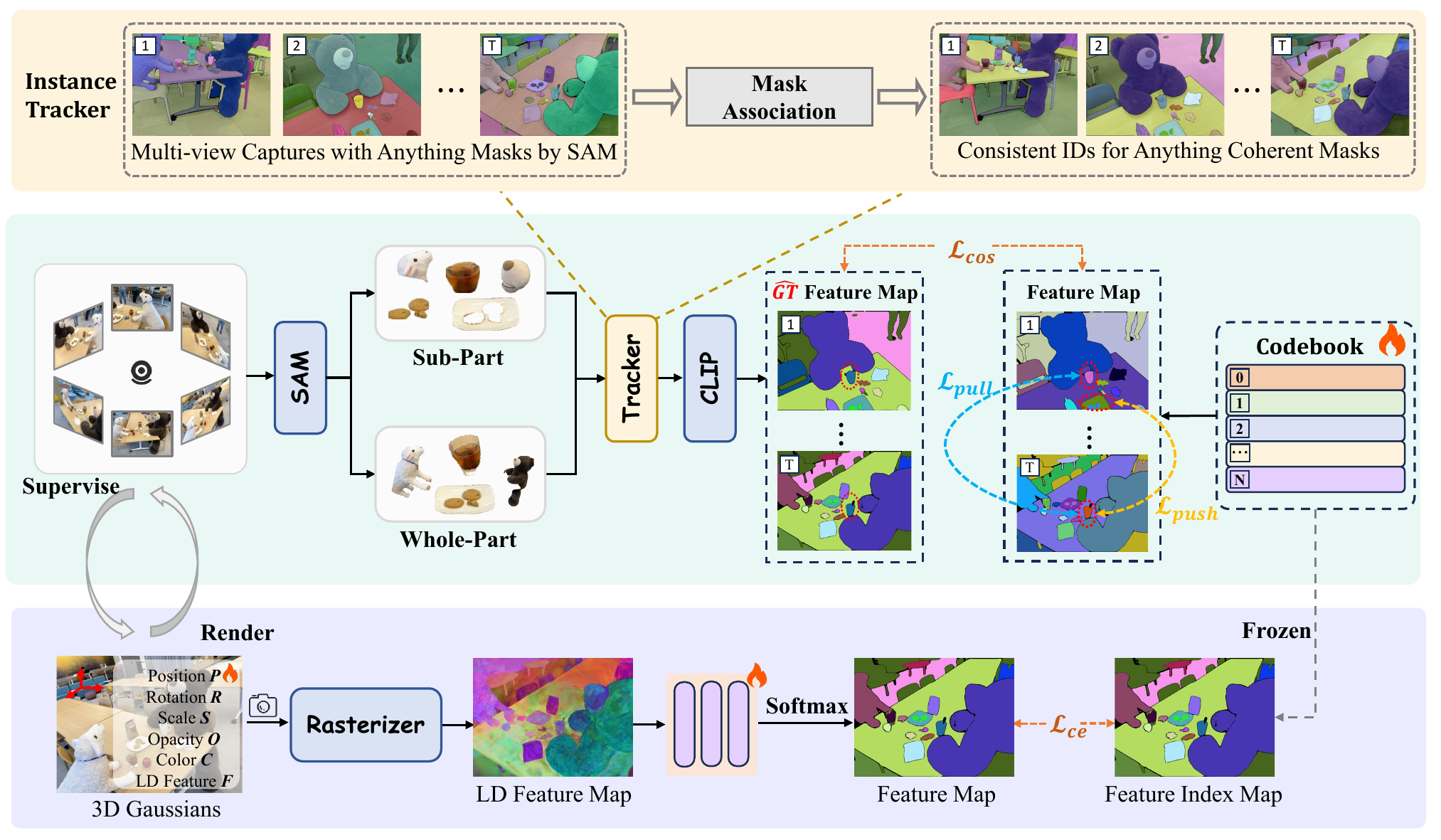}
\vspace{-0.3cm}
\caption{The framework of our CCL-LGS. Top: Instance tracker responsible for mask association. Middle: CCL module that constructs consistent 2D semantic supervision. For multi-view scene images, we first extract dual-scale masks, then perform mask association through the tracker and extract semantics using CLIP. The CCL module subsequently enhances intra-class compactness and inter-class distinctiveness in the categorized semantic features, effectively mitigating ambiguities caused by incomplete or noisy masks. Bottom: The optimization process of the 3D Gaussian semantic field. For each training view, low-dimensional (LD) features from 3D Gaussians are rendered into 2D maps. The optimization is driven by supervision using cross-entropy loss.}
\label{fig:pipeline}
\end{figure*}

In this section, we present our proposed framework, CCL-LGS, for view-consistent 3D semantic reconstruction. As illustrated in Fig.~\ref{fig:pipeline}, we first extract two-level semantic features from multi-view images (Sec.~\ref{sec:feature extraction}), then perform mask association and contrastive codebook learning to organize and refine these features (Sec.~\ref{sec:ccl}), and finally integrate the semantic information into the 3D Gaussian semantic field (Sec.~\ref{sec:semantic_field}).

\subsection{Preliminary: 3DGS with Language Features}
\label{sec:3DGS}

3D Gaussian Splatting utilizes a set of 3D Gaussians, represented as Gaussian ellipsoids, to model a scene. Each 3D Gaussian is parameterized by its centroid $p \in \mathbb{R}^3$, its covariance, which is defined by a rotational quaternion $r \in \mathbb{R}^4$ and a scaling factor $s \in \mathbb{R}^3$. To enable fast $\alpha$-blending for rendering, each Gaussian is also associated with an opacity value $\alpha \in \mathbb{R}$ and a color vector $c$, represented in the three degrees of spherical harmonics (SH) coefficients. To further enhance scene representation, Gaussian splatting models have been extended to incorporate dense language-embedding information~\cite{langsplat,legaussians,zhou2024feature}. This is achieved by introducing a low-dimensional (LD) language feature vector $f \in \mathbb{R}^{d_f}$ for each Gaussian. In summary, the learnable parameters of the $i$-th 3D Gaussian are represented as: $
\theta_i = \{ p_i, r_i, s_i, \alpha_i, c_i, f_i \}.$

Images are rasterized by splatting Gaussians onto each pixel $v$ in the scene and performing $\alpha$-blending to compute the final color $C(v)$, as defined by:
\begin{equation}
C(v) = \sum_{i \in \mathcal{N}} c_i \alpha_i \prod_{j=1}^{i-1} (1 - \alpha_j),
\end{equation}
where $\mathcal{N}$ denotes the set of Gaussians contributing to the tile. Similarly, the predicted semantic feature at pixel $v$ is rendered as:
\begin{equation}
\hat{F}(v) = \sum_{i \in \mathcal{N}} f_i \alpha_i \prod_{j=1}^{i-1} (1 - \alpha_j).
\end{equation}

\subsection{Two-Level Semantic Feature Extraction}
\label{sec:feature extraction}

Existing approaches for 3D semantic understanding that incorporate language embeddings~\cite{lerf,legaussians,goi,langsplat} have demonstrated promising performance. However, some methods~\cite{lerf, legaussians} rely on multi-scale patch averaging for pixel-level semantic feature extraction, which often leads to blurred boundaries. In contrast, GOI’s single-scale mechanism struggles to resolve semantic ambiguities effectively. Although LangSplat~\cite{langsplat} extracts object-level features with clear boundaries by generating masks for subparts, parts, and whole objects, its dependence on multiple models increases data processing and training times, and it may introduce scale errors due to CLIP ambiguities and potential information loss during training. To address these challenges, our approach harnesses SAM's robust segmentation capabilities to achieve precise boundaries while effectively consolidating multi-scale semantic information within a unified framework.

In our method, a uniform 32$\times$32 point prompt is provided to SAM to generate three types of masks corresponding to the semantic scales of subparts, parts, and whole objects. Recognizing that different points may yield conflicting scale assignments (for example, a subpart for one point might be considered a part for another), we merge the subpart and part masks as well as the whole and part masks to obtain two aggregated sets of masks, denoted as \(M^{sp}_o\) and \(M^{wp}_o\). A filtering procedure is then applied to eliminate redundant masks based on predicted IoU scores, stability metrics, and overlap rates, with the final non-overlapping masks \(M_i^l\) (where \(l \in \{sp, wp\}\) and \(i\) indexes the masks) selected based on the product of the IoU and stability scores. For clarity, we omit the superscript \( l \) in all subsequent expressions (e.g., \( M_i \) instead of \( M_i^l \)), with the implicit understanding that operations are independently applied to each scale \( l \). After obtaining segmentation maps, pixel-aligned language embeddings can be extracted using CLIP. For each pixel $v$, its semantic feature $F_i(v)$ can be expressed as:
\begin{equation}
F_i(v) = \text{CLIP}(I_t \odot M_i(v)), \label{supervised_f} 
\end{equation}
where $I_t$ is the input image, and $M_i(v)$ denotes the mask region to which pixel $v$ belongs.

The design of this module is motivated by the need to balance computational efficiency with multi-scale semantic precision. By integrating mask generation and feature extraction within a unified framework, our approach reduces computational overhead while ensuring high semantic accuracy and precise boundary delineation.

\subsection{Contrastive Codebook Learning}
\label{sec:ccl}

Having obtained a set of high-dimensional features, recent methods~\cite{langsplat,legaussians,3dvlgs,goi} typically use an autoencoder or codebook to obtain low-dimensional semantic features. These low-dimensional features are then used to supervise the low-dimensional semantic encoding stored in 3D Gaussians. However, due to occlusion, image blur, and view-dependent variations, applying CLIP directly to imperfect masks results in inconsistent semantic features, ultimately affecting the quality of 3D semantic field reconstruction.

To mitigate the limitations of directly using features derived from imperfect masks, we introduce a codebook-based contrastive learning approach. This approach consists of two key steps: (1) mask association via IoU matching and (2) applying contrastive losses to improve feature representation. 

The first step corresponds to the ``Mask Association" module at the top of Fig.~\ref{fig:pipeline}. Specifically, we first propagate the $K$ masks from the first frame to all frames using SAM2~\cite{sam2}. For the $t$-th frame, the propagated $K$ masks are compared with all masks $M_i$ in that frame using IoU. If the maximum IoU between a propagated mask and a mask $M_i$ exceeds 0.5, then $M_i$ is assigned the category of the matching propagated mask; otherwise, it is assigned the category $-1$ to indicate an unmatched mask. Hence, each mask $M_i$ is assigned a label $y_i \in \{1, 2, \dots, K, -1\}$.

In the second step, a codebook $T = \{T_j\}_{j=1}^{N}$ is constructed, where each prototype $T_j \in \mathbb{R}^d$ is learned during training. Note that $N$ is independent of the $K+1$ mask categories. Specifically, $N$ represents a fixed capacity for scene-specific feature learning, while $K$ refers to the number of object categories observed in the current scene subset, which may be incomplete due to limited observations. The codebook serves as latent feature prototypes that structure the feature space for contrastive learning. Contrastive losses are then applied to encourage the alignment of features with the same category (pull loss) and the separation of features with different categories (push loss), as shown in the middle of Fig.~\ref{fig:pipeline}.

For each feature \(F_i\), we search for the most similar prototype in \(T\) based on cosine similarity:
\begin{equation}
j^* = \underset{j}{\text{argmax}} \; \cos(F_i, T_j), \label{argmax}
\end{equation}
where \(\cos(F_i, T_j)\) denotes the cosine similarity between the feature \(F_i\) and the prototype \(T_j\). To encourage \(F_i\) to match well with its closest prototype, we define a matching loss:
\begin{equation}
L_{\text{max}} = 1 - \cos(F_i, T_{j^*}).
\end{equation}

Contrastive losses are then applied based on the assigned mask labels. For two features \(F_i\) and \(F_k\) with corresponding mask labels \(y_i\) and \(y_k\):

If \(y_i = y_k\) and \(y_i \neq -1\), we apply a pull loss to pull their associated descriptors closer via their codebook projections:
\begin{equation}
  L_{\text{pull}} = 1 - \cos(T_{j_i}, T_{j_k}),
\end{equation}
  where \(T_{j_i}\) and \(T_{j_k}\) are the prototypes selected for \(F_i\) and \(F_k\), respectively.

If \(y_i \neq y_k\) and \(y_i, y_k \neq -1\), we apply a push loss to force their codebook representations apart:
\begin{equation}
  L_{\text{push}} = \text{ReLU}(\cos(T_{j_i}, T_{j_k}) - m),
\end{equation}
  where \(m > 0\) is a predefined margin.

For features with \(y_i = -1\) (unmatched masks), neither pull nor push loss is applied.
Finally, the total loss is formulated as a weighted sum:
\begin{equation}
L = L_{\text{max}} + \lambda_{\text{pull}} L_{\text{pull}} + \lambda_{\text{push}} L_{\text{push}},
\end{equation}
with weighting factors \(\lambda_{\text{pull}},\) and \(\lambda_{\text{push}}\) controlling the contribution of each component.

This framework design has two advantages. First, compared to autoencoders, similar features in the feature space are implicitly constrained to the same table entry, resulting in stronger consistency constraints. Second, through contrastive learning losses, the framework ensures that features corresponding to the same mask category become well clustered, while those from different mask categories are effectively separated. This improves the semantic consistency of imperfect masks and increases the distinction between semantics, resulting in better 2D supervised representations.
\subsection{3D Gaussian Semantic Field}
\label{sec:semantic_field}
Given a trained codebook \(T = \{T_j\}_{j=1}^N\), we construct the 3D Gaussian semantic field as illustrated at the bottom of Fig.~\ref{fig:pipeline}, by converting per-pixel semantic features into discrete indices and aligning these indices with the outputs of 3D Gaussian Splatting.

Specifically, for each pixel \(v\), its semantic feature \(F_i(v)\) is obtained via Eq.~\eqref{supervised_f}. The index \(j^*\) is assigned to \(v\) through Eq.~\eqref{argmax}, generating a semantic index map \(\mathcal{M} \in \mathbb{R}^{H\times W}\), where \(H\) and \(W\) denote the image height and width, respectively. Subsequently, a feature map \(\hat{F} \in \mathbb{R}^{H\times W\times d_f}\) is rendered via rasterization and alpha blending (see Sec.~\ref{sec:3DGS}). This feature map is processed by a lightweight MLP decoder followed by a softmax layer to produce a semantic feature distribution \(\hat{\mathcal{M}} \in \mathbb{R}^{H\times W\times N}\), where \(N\) corresponds to the number of codebook categories. To jointly optimize the semantic features of 3D Gaussians and the parameters of the MLP decoder, we minimize the cross-entropy loss:
\begin{equation}
  \mathcal{L}_{\text{CE}} = \text{CE}(\hat{\mathcal{M}}, \mathcal{M}),
\end{equation}

At inference, each pixel retrieves its prototype \(T_{\hat{\mathcal{M}}(v)}\) to form the refined semantic feature \(\tilde{F}(v)\). Given a text query \(\tau\), we compute its embedding \(\varphi(\tau)\) via the text encoder of the vision-language model to compute the relevance map.

\begin{equation}
    p(\tau \mid v) = \frac{\exp\left(\frac{\tilde{F}(v) \cdot \varphi(\tau)}{\lVert \tilde{F}(v) \rVert \lVert \varphi(\tau) \rVert}\right)}{\sum_{s \in \mathcal{T}} \exp\left(\frac{\tilde{F}(v) \cdot \varphi(s)}{\lVert \tilde{F}(v) \rVert \lVert \varphi(s) \rVert}\right)}.
\end{equation}
Thresholding \(p(\tau\mid v)\) yields a semantic segmentation mask for the queried concept.

\section{Experiments}
{\flushleft\textbf{Dataset.}}
To assess the effectiveness of our approach, we conduct experiments on two benchmark datasets using the mean Intersection over Union (mIoU) metric. The first dataset, LERF~\cite{lerf}, is captured using the Polycam application on an iPhone and features four challenging indoor scenes: Ramen, Figurines, Teatime, and Waldo Kitchen. These scenes are annotated with pixel-accurate ground truth masks for textual queries, as provided by the LangSplat~\cite{langsplat}. The dataset's real-world imaging conditions, including severe occlusions and motion blur, make it particularly suited for testing segmentation robustness in complex environments. The second dataset, 3D-OVS~\cite{liu2023weakly}, consists of long-tail objects set against diverse backgrounds. For our evaluation, we focus on four specific scenes: Bed, Bench, Lawn, and Sofa. Note that the Room scene contains a significant annotation error; thus, we exclude it from quantitative evaluation and provide qualitative results only in the supplementary material.

{\flushleft\textbf{Implementation Details.}}
To extract semantic features of each image, we utilize the SAM ViT-H model~\cite{sam} alongside the OpenCLIP ViT-B/16 model~\cite{clip}.
For the contrastive loss coefficients, we set $\lambda_{\text{pull}} = \lambda_{\text{push}} = 0.25$ and use a margin $m$ of 0.7 to ensure adequate separation between feature clusters.
For each scene, we jointly train the 3D Gaussians for both color and semantic features by setting $d_c=3$ and $d_f=8$. The training is performed over 30,000 iterations using the Adam optimizer~\cite{Kingma_Ba_2014}, with a learning rate of 0.001 and beta parameters set to $(0.9, 0.999)$. 
{\flushleft\textbf{Baseline.}}
For a fair comparison, we select the latest works on open-vocabulary 3D scene understanding: Feature-3DGS\cite{zhou2024feature}, LEGaussians\cite{legaussians}, LangSplat\cite{langsplat}, GS-Grouping\cite{gaussiangrouping}, GOI\cite{goi}, and 3D VL-GS\cite{3dvlgs}.

\subsection{Experiments on LERF}
\begin{figure*}[t]
\centering
\includegraphics[width=1.0\textwidth]{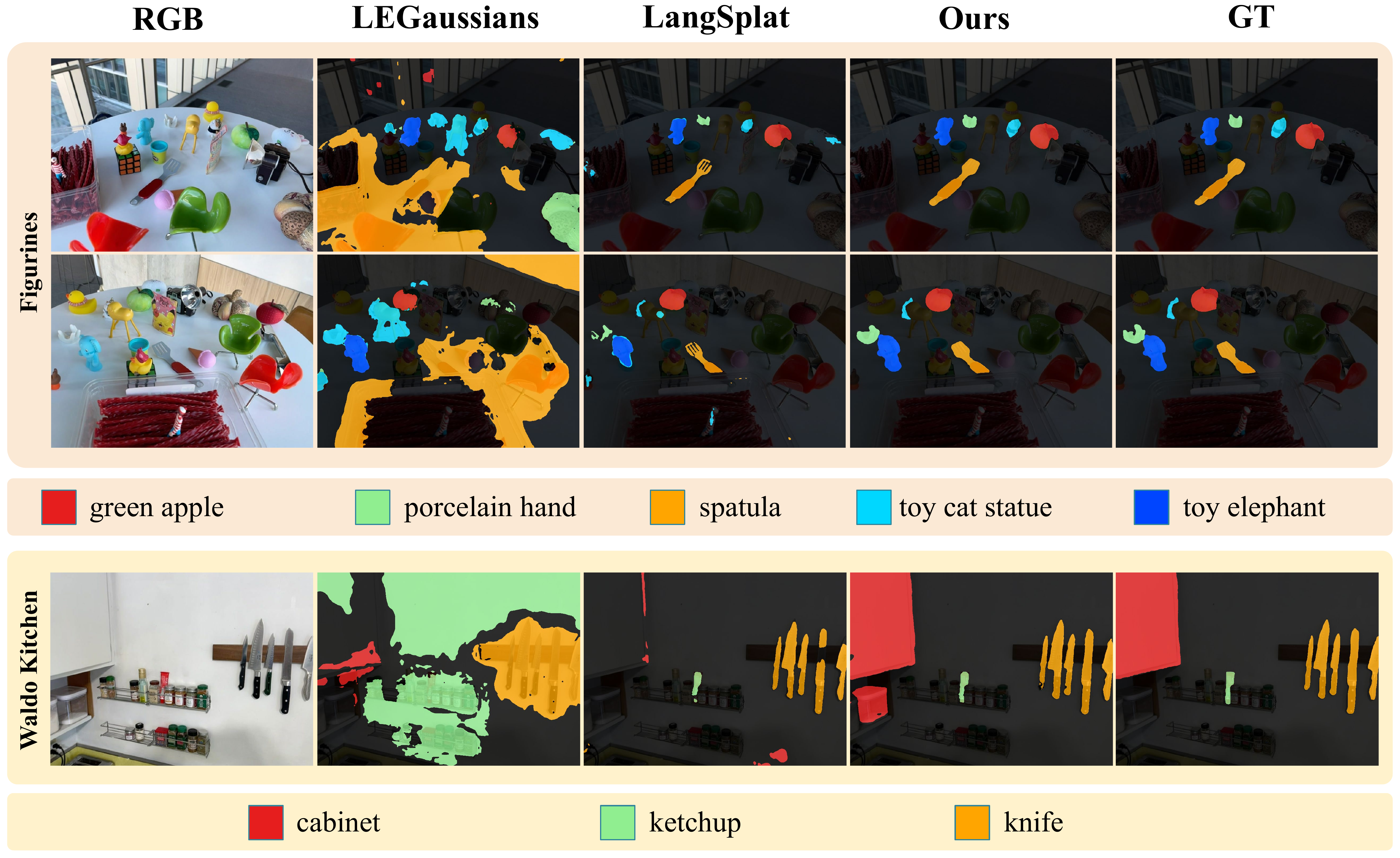}
\vspace{-0.4cm}
\caption{Segmentation results on the figurines (top) and kitchen (bottom) scenes. Our method achieves consistent multi-view segmentation and accurately captures challenging objects like the cabinet, outperforming prior approaches.}
\label{fig:lerf}
\vspace{-0.3cm}
\end{figure*}

{\flushleft\textbf{Quantitative Results.}}
We first compare our method with existing SOTA methods on LERF dataset. As shown in Tab.~\ref{tab:LERF}. We observed that our method achieved an IoU result of 65.6 in 3D semantic segmentation, ranking either first or second across all four scenes, outperforming the state-of-the-art 3D Vision-Language GS by 3.6. This illustrates the superiority of our proposed CCL-LGS method.
{\flushleft\textbf{Visualization Results.}}
Fig.~\ref{fig:lerf} illustrates segmentation results for two scenes: figurines (top) and kitchen (bottom). In the figurines scene, we compare how each method segments the same object across two different viewpoints, revealing that competing approaches often exhibit inconsistent segmentations. In contrast, our method provides stable and accurate results, even under viewpoint changes. In the kitchen scene, we specifically focus on the cabinet, a challenging object that other methods frequently fail to segment correctly. Our CCL-LGS framework effectively captures its boundaries, demonstrating both superior cross-view consistency and strong performance on difficult categories.

\begin{table}[t]
    \centering
    \resizebox{1.0\columnwidth}{!}{
    \begin{tabular}{lccccc}
        \toprule
        Method & Ramen  & Figurines  & Teatime  &  Waldo Kitchen & Avg.  \\  \hline
        Feature-3DGS & 43.7 & 40.5 & 58.8 & 39.6 & 45.7  \\ 
        LEGaussians & 46.0 & 40.8 & 60.3 & 39.4 & 46.6  \\ 
        LangSplat & 51.2 & 44.7 & 65.1 & 44.5 & 51.4  \\ 
        GS-Grouping & 45.5 & 40.0 & 60.9 & 38.7 & 46.3  \\ 
        GOI & 52.6 & 44.5 & 63.7 & 41.4 & 50.6  \\ 
        3D VL-GS & \underline{61.4} & 58.1 & \textbf{73.5} & 54.8 & \underline{62.0}  \\ \hline
        Ours & \textbf{62.3} & \textbf{61.2} & \underline{71.8} & \textbf{67.1} & \textbf{65.6}  \\ 
        \bottomrule
    \end{tabular}}
        \caption{Quantitative experiments results on LERF dataset. The best result is bolded, and the second-best result is underlined.}
    \label{tab:LERF}
    \vspace{-0.2cm}
\end{table}

\begin{table}[t]
    \centering

    \resizebox{1.0\columnwidth}{!}{
    \begin{tabular}{lccccc}
        \toprule
        Method & Ramen  & Figurines  & Teatime  &  Waldo Kitchen & Avg.  \\  \hline
        Baseline & 46.8 & 57.1 & 60.8 & 61.0 & 56.4  \\ 
        baseline(w/ pull loss)& 48.0 & 58.0 & 70.1 & 62.0 & 59.5  \\ 
        baseline(w/ push loss) & 55.1 & 61.0 & 66.0 & 59.3 & 60.4  \\ 
        Ours & \textbf{62.3} & \textbf{61.2} & \textbf{71.8} & \textbf{67.1} & \textbf{65.6}  \\ 
        \bottomrule
    \end{tabular}}
    \caption{Ablation study on LERF dataset.}
    \label{tab:ablation}
    \vspace{-0.2cm}
\end{table}

\begin{figure}[t]
\centering
\includegraphics[width=1\linewidth]{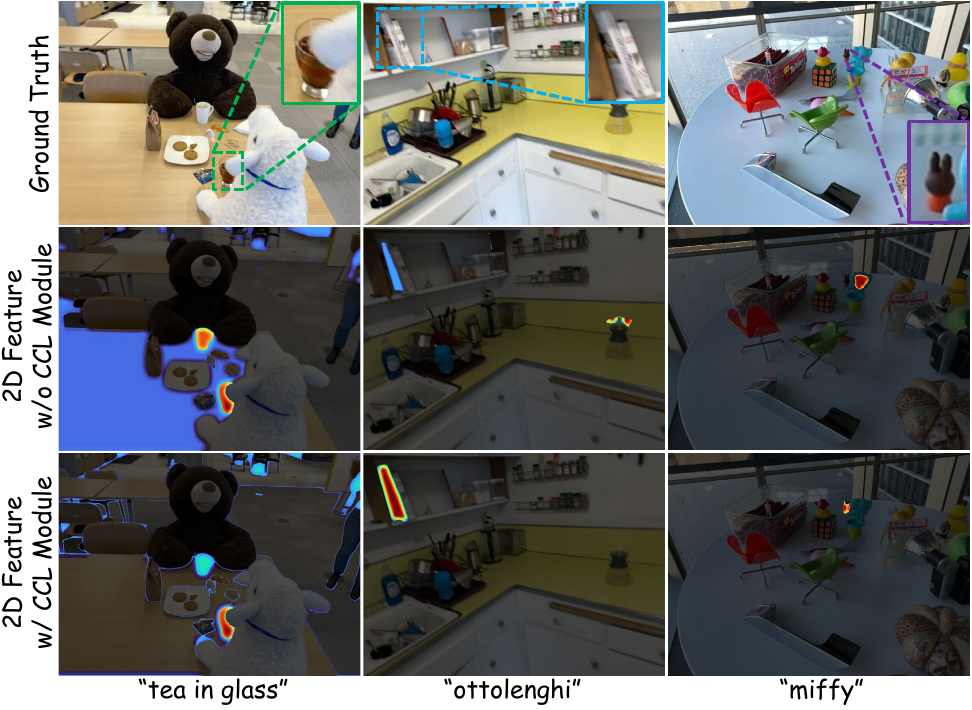}
\vspace{-0.4cm}
\caption{Qualitative comparison of 2D feature maps with and without CCL module. }
\label{fig:ab2}
\vspace{-0.4cm}
\end{figure}

\begin{figure*}[tb]
\centering
\includegraphics[width=1.0\textwidth]{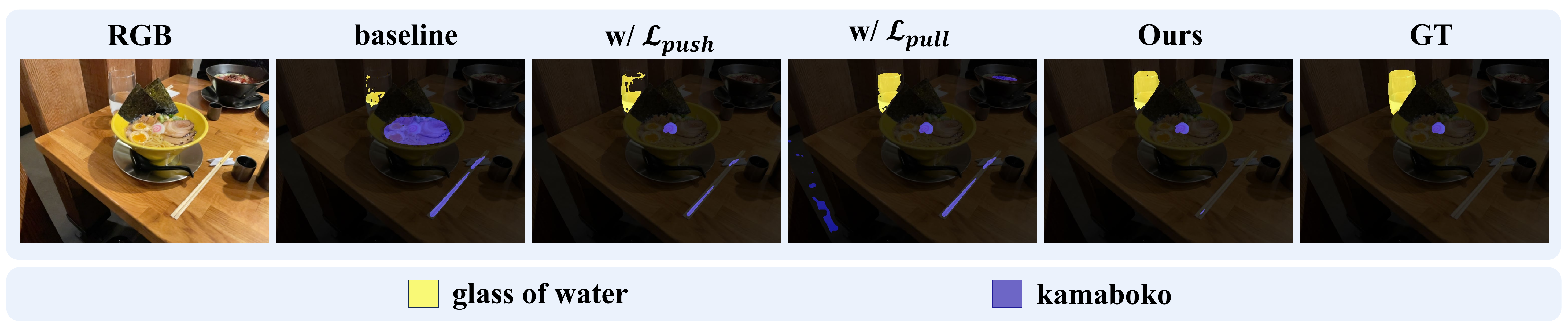}
\vspace{-0.3cm}
\caption{Qualitative comparison of different loss configurations. The pull loss improves intra-class consistency (e.g., for ``glass of water"), while the push loss reduces false activations (e.g., around the ``kamaboko"). The full model effectively combines both.}
\vspace{-0.4cm}
\label{fig:ab1}
\end{figure*}

\begin{figure*}[t]
\centering
\includegraphics[width=1.0\textwidth]{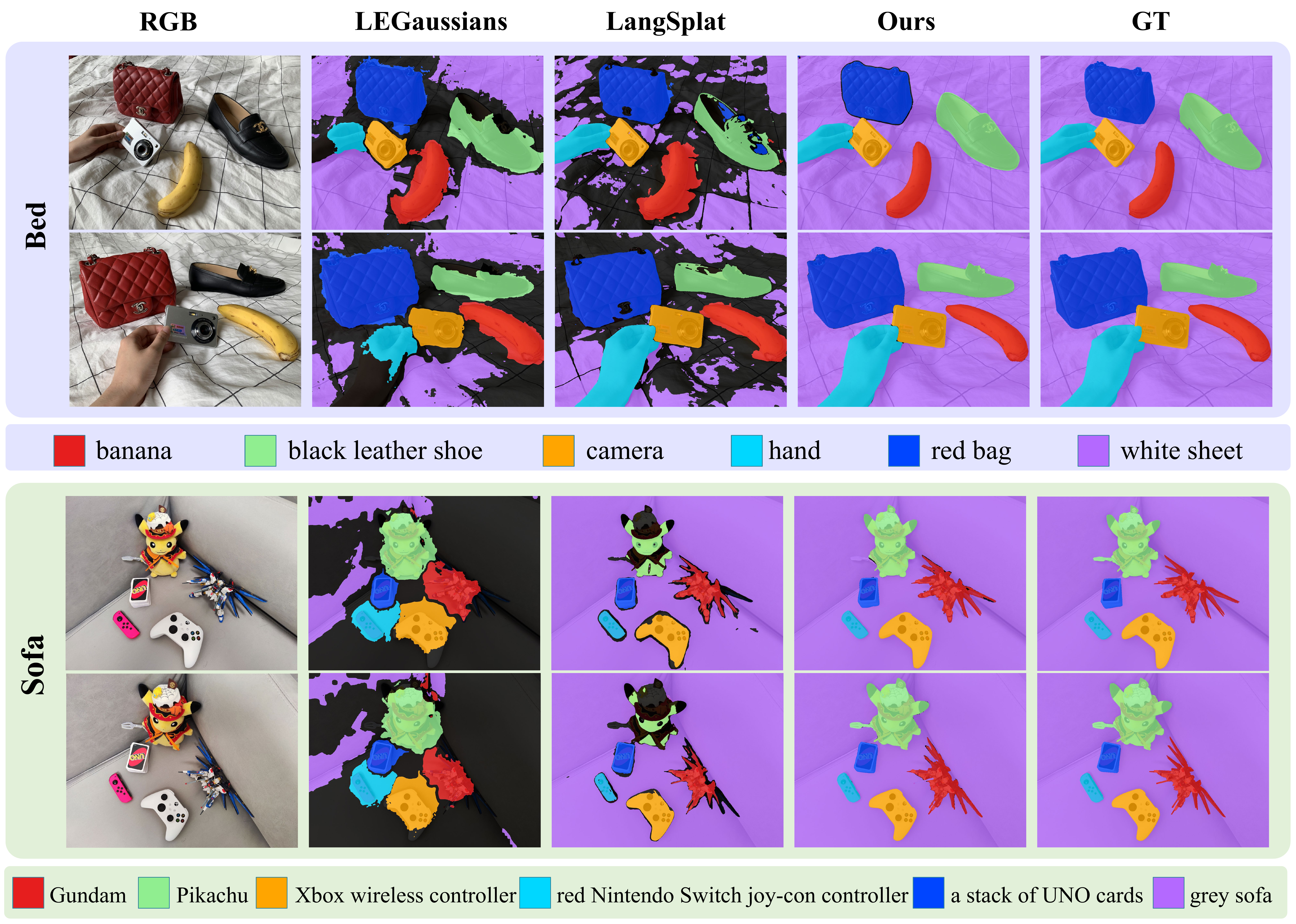}
\vspace{-0.4cm}
\caption{Qualitative comparison on 3D-OVS dataset. In these two scenes, our method clearly outperforms the other two methods.}
\label{fig:3dovs}
\vspace{-0.4cm}
\end{figure*}

{\flushleft\textbf{Ablation study.}}
To validate the effectiveness of our Contrastive Codebook Learning (CCL) module, we conduct experiments, including visual analysis of 2D supervision features and ablation studies on 3D semantic segmentation. As shown in Fig.~\ref{fig:ab2}, features refined by CCL exhibit higher spatial precision and stronger activation on semantically relevant regions, eliminating scattered and ambiguous responses in the baseline. This demonstrates that CCL improves alignment between extracted features and semantic ground truth during 2D supervision.

We further perform ablation studies to analyze the impact of individual loss components on 3D semantic understanding. Evaluations are conducted on four LERF scenes. The baseline uses codebook-based feature compression to produce 2D semantic supervision for 3D segmentation. We compare three variants: baseline with pull loss, baseline with push loss, and the full model with both. As shown in Tab.~\ref{tab:ablation}, both losses are essential for optimal performance—removing either causes noticeable degradation, though all variants still surpass the baseline.

Qualitative results in Fig.~\ref{fig:ab1} further show the complementary roles of the two losses. The pull loss improves intra-class consistency, especially for occluded or partly visible objects like the ``glass of water”, while the push loss reduces false activations in confusing regions, such as near the ``kamaboko”. Combining both ensures robust, discriminative 3D semantic segmentation in challenging scenes.

\begin{table}[t]
    \centering
    \resizebox{1.0\columnwidth}{!}{
    \begin{tabular}{lccccc}
        \toprule
        Method & Bed & Bench & Sofa & Lawn & Avg.  \\  \hline
        Feature-3DGS & 83.5 & 90.7 & 86.9 & 93.4 & 88.6  \\ 
        LEGaussians & 84.9 & 91.1 & 87.8 & 92.5 & 89.1  \\ 
        LangSplat & 92.5 & 94.2 & 90.0 & \underline{96.1} & 93.2  \\ 
        GS-Grouping & 83.0 & 91.5 & 87.3 & 90.6 & 88.1  \\ 
        GOI & 89.4 & 92.8 & 85.6 & 94.1 & 90.5  \\ 
        3D VL-GS & \underline{96.8} & \textbf{97.3} & \textbf{95.5} & \textbf{97.9} & \textbf{96.9}  \\  \hline
        Ours & \textbf{97.3} & \underline{95.0} & \underline{92.3} & \underline{96.1} & \underline{95.2}  \\ 
        \bottomrule
    \end{tabular}}
        \caption{Quantitative experiments results on 3D-OVS dataset. The best result is bolded, and the second-best result is underlined.}
    \label{tab:3D-OVS}
    \vspace{-0.4cm}
\end{table}

\subsection{Experiments on 3D-OVS}
{\flushleft\textbf{Quantitative Results.}}
We compare our method with existing state-of-the-art approaches on the 3D-OVS dataset, as shown in Tab.~\ref{tab:3D-OVS}. While our approach achieves comparable performance, it underperforms 3D VL-GS. We attribute this to the nature of the 3D-OVS dataset, which is relatively small and simple, with minimal occlusion and blur—conditions under which the data enrichment strategy employed by 3D VL-GS offers more noticeable benefits.

{\flushleft\textbf{Qualitative Results.}}
We present our qualitative experimental results in Figure \ref{fig:3dovs}. Since LEGaussians is a patch-based method, it learns over-smoothed features and fails to capture sharp object boundaries. LangSplat adopts a three-scale architecture but does not enhance multiview features as our method does. In our experiments, we observed that it sometimes selects inappropriate scales for certain objects, leading to suboptimal segmentation performance. Among the compared methods, our method produces the most accurate segmentation maps, further demonstrating the effectiveness of our CCL-LGS.

\section{Conclusion}
In this paper, we present CCL-LGS, a novel method to build 3D language fields enabling accurate, efficient open-vocabulary queries. We address an overlooked challenge: directly applying CLIP to imperfect masks yields inconsistent semantic features, introducing artifacts in 3D reconstruction. To solve this, we propose a dedicated CCL module that creates compact, distinct features for more reliable supervision. Extensive experiments show state-of-the-art performance. Limitations remain due to inherent capabilities of SAM and SAM2, as imperfect masks still affect results. Future work will refine masks for greater robustness.

{
    \small
    \bibliographystyle{ieeenat_fullname}
    \bibliography{main}
}
\clearpage
\appendix
% \clearpage
% \setcounter{page}{1}
\maketitlesupplementary

\section{Qualitative Results on the Room Scene}
\label{sec:room_qualitative}

\begin{figure}[t]
\centering
\includegraphics[width=1\linewidth]{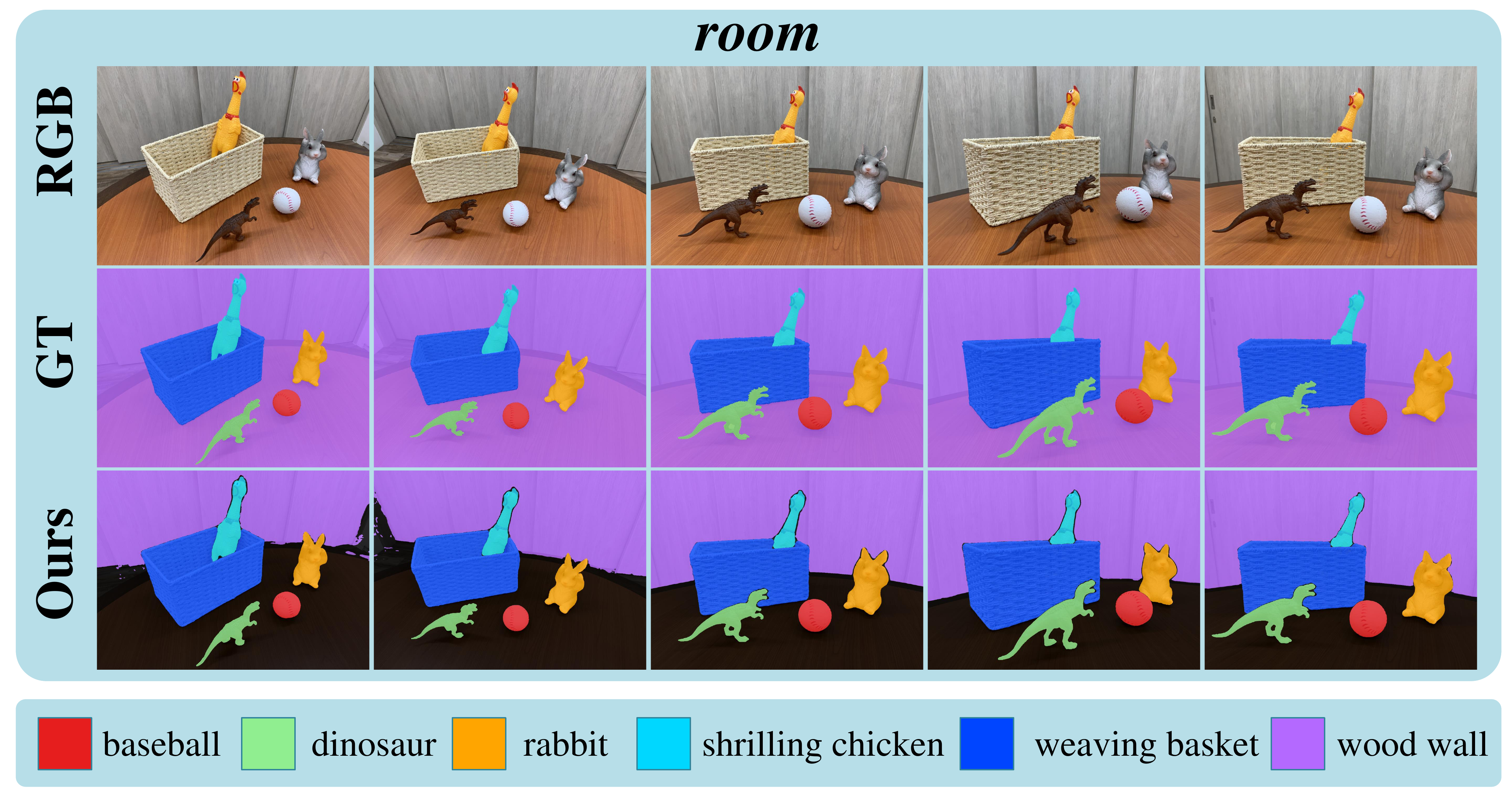}
\vspace{-0.4cm}
\caption{Qualitative results on the Room scene. The ``wood wall" category contains obvious annotation errors.}
\vspace{-0.4cm}
\label{fig:room}
\end{figure}

As mentioned in the main paper, the Room scene in the 3D-OVS dataset includes annotation errors specifically in the ``wood wall" region. This mislabeling affects the reliability of quantitative evaluation. Therefore, we present qualitative results in Fig.~\ref{fig:room}.

\section{Efficiency Analysis}
\label{sec:efficiency}

We conducted additional experiments on the Ramen scene from the LERF dataset using an Intel i7-14700KF CPU and an NVIDIA RTX 4090 GPU. The results are summarized in Tab.~\ref{tab:efficiency}. The rendering speed (FPS) is measured by rendering the images with language features at a consistent resolution. Results show that our method achieves a favorable balance between multi-scale segmentation accuracy and efficiency under constrained GPU memory conditions.

\section{Scale Number Analysis}

To explore the impact of scale granularity on semantic segmentation accuracy, we compare our two-scale design with a baseline that merges all masks from three scales (subparts, parts, and whole objects) into a single set, as shown in Tab.~\ref{tab:scale_ab}. This single-scale baseline simplifies processing but sacrifices scale-specific representation.

Although using all three separate scales may provide marginal performance gains, we found that it leads to prohibitive GPU memory usage, system memory consumption, and preprocessing time due to the overwhelming number of fine-grained masks generated at the subpart level. Given these limitations, we do not include full three-scale experiments.

Our method merges subpart with part masks and whole with part masks, producing two non-overlapping, semantically meaningful sets. This strategy reduces redundancy and computational overhead while preserving scale-aware distinctions.

\section{Language-based 3D Interaction Capability}

Although our main paper emphasizes 2D supervision for semantic learning, our method indeed supports direct 3D interaction via language queries. Our codebook-based approach allows users to perform language-based querying and editing directly in 3D space. 

\begin{figure}[t]
\centering
\includegraphics[width=1\linewidth]{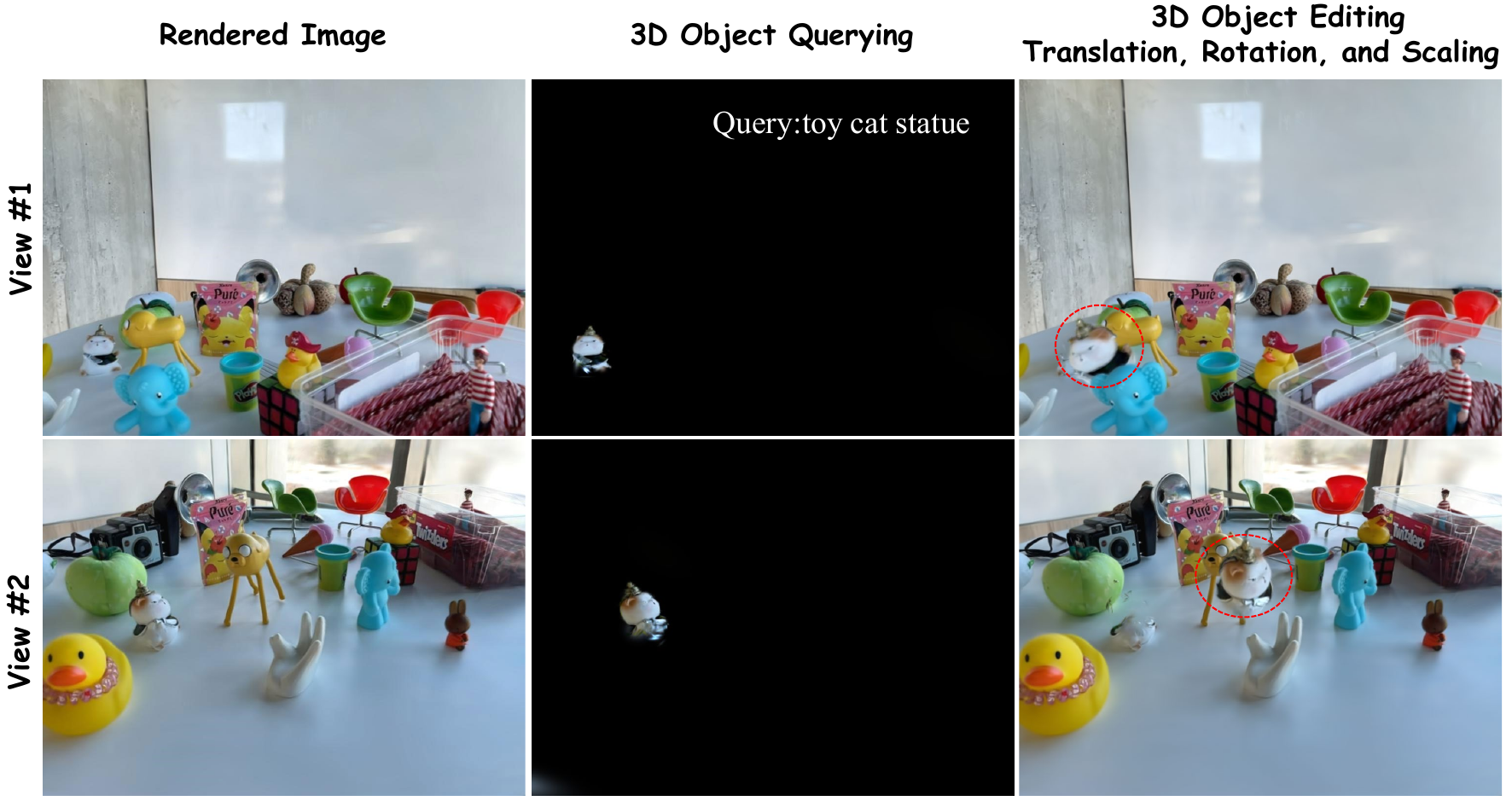}
\vspace{-0.4cm}
\caption{Examples of language-based 3D interaction and editing enabled by our method.}
\label{fig:3d1}
\vspace{-0.4cm}
\end{figure}

Specifically, given a language query, we first match the query embedding with the codebook to select relevant semantic categories. Then, without any rasterization or alpha blending, we directly classify the semantic features of each 3D Gaussian using a lightweight linear classifier. Based on the predicted category distribution, we can identify Gaussians that correspond to the selected categories. This enables us to identify Gaussians in the 3D space that are semantically aligned with the input query. Once the relevant Gaussians are localized, we can directly perform various interaction operations on them, thereby enabling intuitive and interpretable 3D editing driven by natural language. Examples of such 3D interactions are illustrated in Fig.~\ref{fig:3d1}.

\begin{table}[t]
    \centering
    \resizebox{1.0\columnwidth}{!}{
    \begin{tabular}{lcccccc}
        \toprule
        Method  & mIoU & Pre-process & Training & Total & FPS & Memory \\
        \midrule
        LangSplat & 51.2 & 256min  & 36min & 292min & 42 & 7GB+4GB \\
        LEGaussians & 46.0 & 2min & 40min & 42min & 65 & 19GB+9GB \\
        Ours & 62.3 & 16min & 74min & 90min & 65 & 11GB+9GB \\
        \bottomrule
    \end{tabular}}
    \caption{Efficiency comparison on the Ramen scene from the LERF dataset. }
\label{tab:efficiency}
\vspace{-0.4cm}
\end{table}

\begin{table}[t]
    \centering
    \resizebox{1.0\columnwidth}{!}{
    \begin{tabular}{lccccc}
        \toprule
        Method & Ramen & Figurines & Teatime & Waldo Kitchen & Avg.  \\ 
        \midrule
        single scale & 44.3 & 57.6 & 70.1 & 57.7 & 57.4  \\ 
        ours & 62.3 & 61.2 & 71.8 & 67.1 & 65.6  \\
        \bottomrule
    \end{tabular}}
    \caption{Comparison of different scale aggregation strategies. }
\label{tab:scale_ab}
\vspace{-0.4cm}
\end{table}

\section{More Results}
\label{sec:additional_viz}

\begin{figure}[h]
\centering
\includegraphics[width=0.95\linewidth]{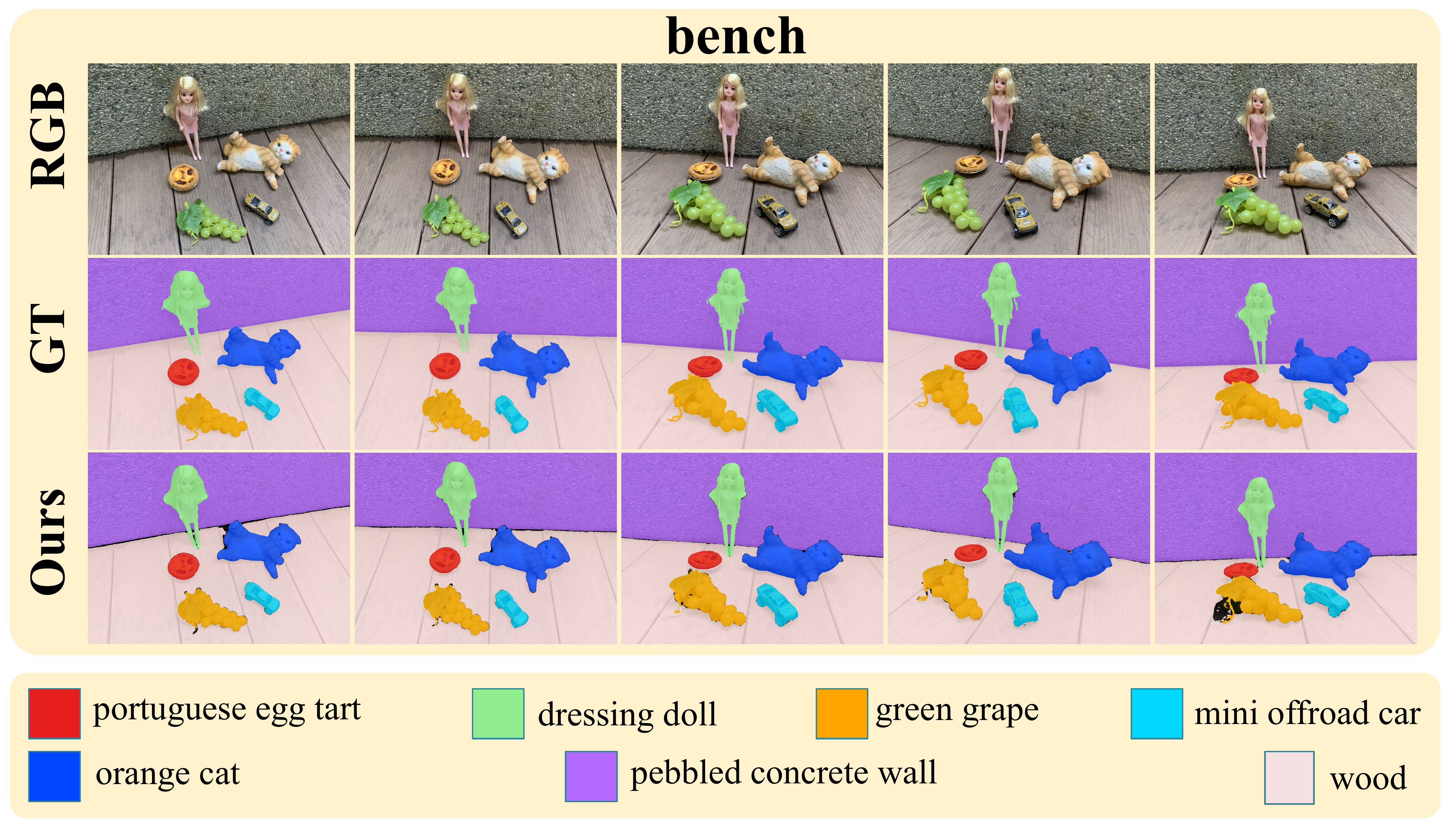}
\caption{Qualitative semantic segmentation results on the Bench scene.}
\label{fig:bench}
\vspace{-0.4cm}
\end{figure}

\begin{figure}[h]
\centering
\includegraphics[width=0.95\linewidth]{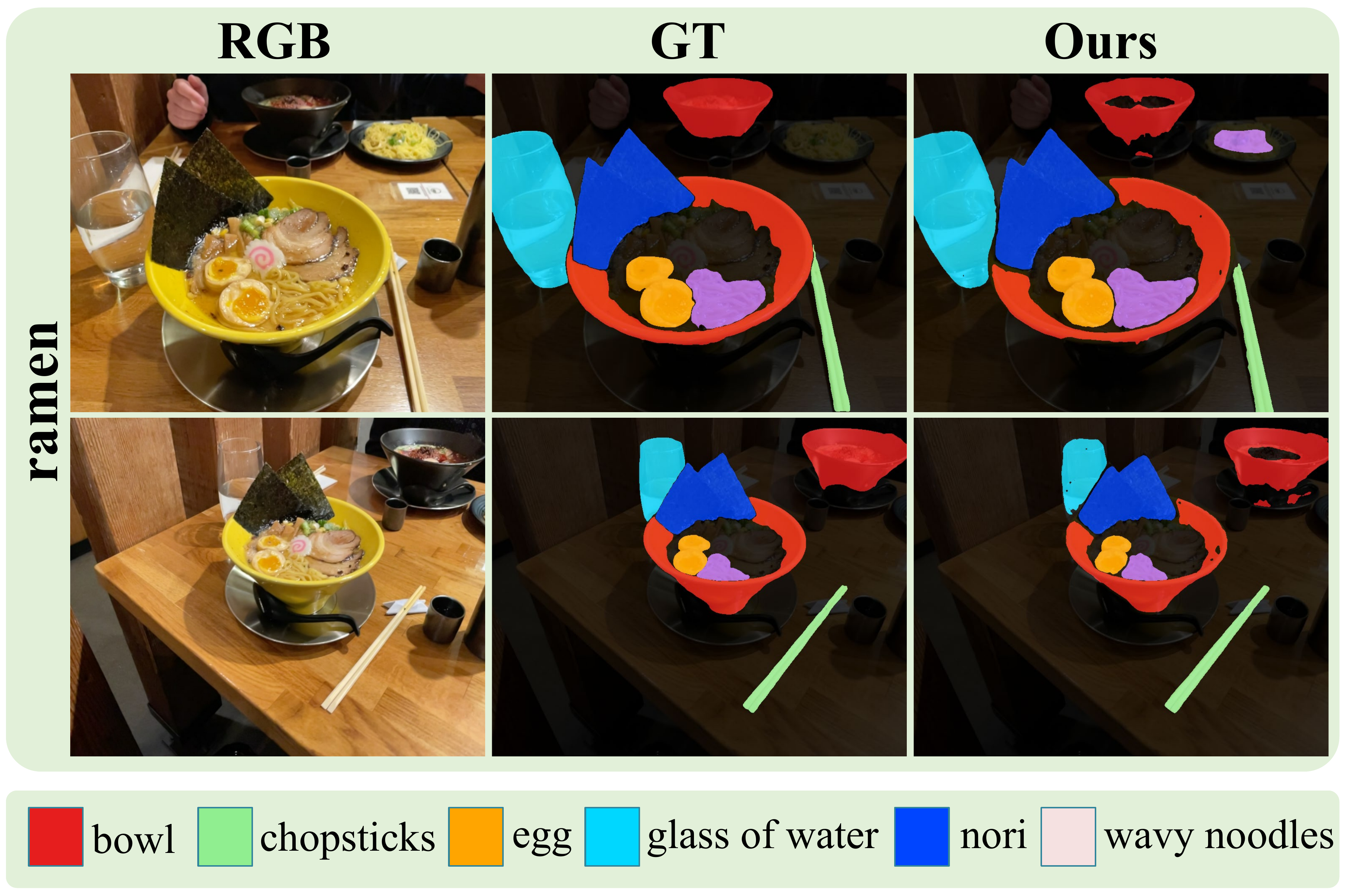}
\caption{Qualitative semantic segmentation results on the Ramen scene.}
\label{fig:ramen}
\vspace{-0.4cm}
\end{figure}

\begin{figure}[h]
\centering
\includegraphics[width=0.95\linewidth]{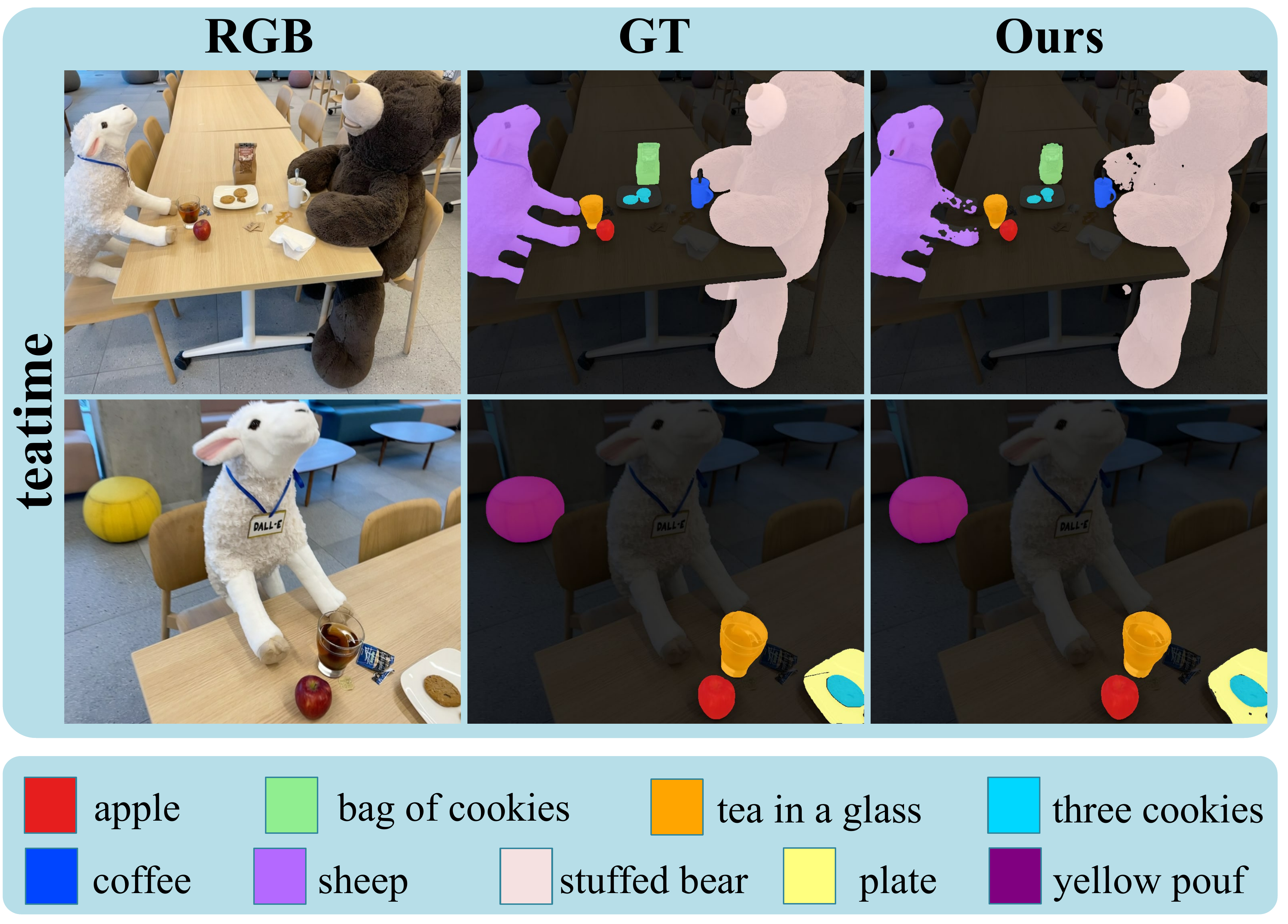}
\caption{Qualitative semantic segmentation results on the Teatime scene.}
\label{fig:teatime}
\end{figure}

Beyond the specific cases discussed, we provide more qualitative visualizations of our method's semantic segmentation results across different scenes in Figs.~\ref{fig:bench}, \ref{fig:ramen}, and \ref{fig:teatime}. 

\end{document}